\documentclass[letterpaper, 10 pt, conference]{ieeeconf}  % Comment this line out if you need a4paper

\IEEEoverridecommandlockouts                              % This command is only needed if 
\usepackage{graphics} % for pdf, bitmapped graphics files
\usepackage{epsfig} % for postscript graphics files
\usepackage{mathptmx} % assumes new font selection scheme installed
\usepackage{times} % assumes new font selection scheme installed
\usepackage{amsmath} % assumes amsmath package installed
\usepackage{amssymb}  % assumes amsmath package installed
\usepackage[T1]{fontenc}
\usepackage{xcolor}

\usepackage[noadjust]{cite}
\usepackage{gensymb}
\usepackage{soul}
\usepackage{url}

\title{\LARGE \bf
On the Fly Robotic-Assisted Medical Instrument Planning and Execution Using Mixed Reality
}

\author{Letian Ai$^{1}$, Yihao Liu$^{1,2,\dagger}$, Mehran Armand$^{1,2,4}$, Amir Kheradmand$^{3,\ddagger}$, and Alejandro Martin-Gomez$^{1,2,\ddagger}$
\thanks{Manuscript submitted Sept 15, 2023. This work was supported by grants from the National Institute of Deafness and Other Communication Disorders (R01DC018815) and National Institute of Biomedical Imaging and Bioengineering (R01EB023939).}
\thanks{$^{1}$ Biomechanical- and Image-Guided Surgical Systems (BIGSS) laboratory within LCSR, Johns Hopkins University, Baltimore, MD, USA
        {\tt\small (lai2@jhu.edu, yliu333@jhu.edu, mehran.armand @jhuapl.edu, alejandro.martin@jhu.edu)}}%
\thanks{$^{2}$ Department of Computer Science, Johns Hopkins University, Baltimore, MD, USA
        {\tt\small }}%
\thanks{$^{3}$ Department of Neurology and Department of Neuroscience, Johns Hopkins School of Medicine, Baltimore, MD, USA
        {\tt\small (akherad1@jhu.edu)}}
\thanks{$^{4}$ Department of Orthopedic Surgery, Johns Hopkins School of Medicine, Baltimore, MD, USA}
\thanks{$^{\dagger}$ Corresponding author}
\thanks{$^{\ddagger}$ Equal contribution}
}

\begin{document}

\maketitle
% \thispagestyle{empty}
% \pagestyle{empty}

%%%%%%%%%%%%%%%%%%%%%%%%%%%%%%%%%%%%%%%%%%%%%%%%%%%%%%%%%%%%%%%%%%%%%%%%%%%%%%%%
\begin{abstract}
Robotic-assisted medical systems (RAMS) have gained significant attention for their advantages in alleviating surgeons' fatigue and improving patients' outcomes. 
These systems comprise a range of human-computer interactions, including medical scene monitoring, anatomical target planning, and robot manipulation. 
However, despite its versatility and effectiveness, RAMS demands expertise in robotics, leading to a high learning cost for the operator.
In this work, we introduce a novel framework using mixed reality technologies to ease the use of RAMS.
%To bridge this gap, this study proposes an innovative mixed reality approach to enhance the RAMS interface, providing informative and user-friendly interactions. 
The proposed framework achieves real-time planning and execution of medical instruments by providing 3D anatomical image overlay, human-robot collision detection, and robot programming interface. These features, integrated with an easy-to-use calibration method for head-mounted display, improve the effectiveness of human-robot interactions. 
%The anatomy data overlay offers immersive and intuitive medical data visualizations, aiding in the identification of anatomical areas of interest and assisting in medical instrument planning. The robot programming method, by defining instrument waypoints and incorporating spatial information from the operator, provides an interface to plan flexible and safe trajectories. Addressing the challenges of aligning rendered graphics with physical objects, a novel method has been implemented to calibrate the mixed reality headset with the external tracking system.
To assess the feasibility of the framework, two medical applications are presented in this work: 1) coil placement during transcranial magnetic stimulation and 2) drill and injector device positioning during femoroplasty. Results from these use cases demonstrate its potential to extend to a wider range of medical scenarios.

\end{abstract}

% \begin{keywords}
% Robotic-assisted medical system, instrument planning and execution, mixed reality
% \end{keywords}
%%%%%%%%%%%%%%%%%%%%%%%%%%%%%%%%%%%%%%%%%%%%%%%%%%%%%%%%%%%%%%%%%%%%%%%%%%%%%%%%
\section{INTRODUCTION}

\begin{figure*}[h!]
\centering 
\includegraphics[width=0.9\linewidth]{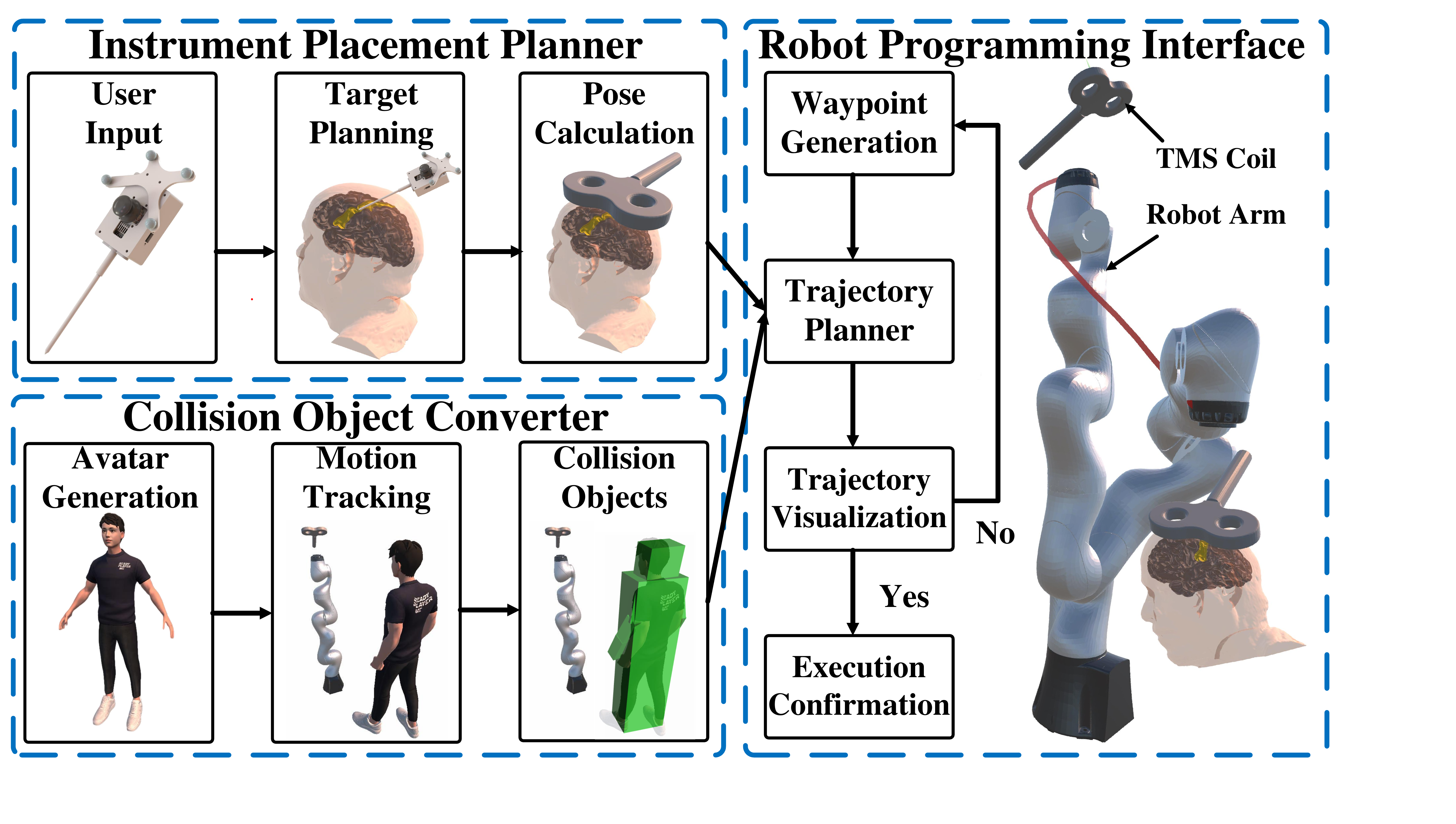}
\caption{System diagram. Arrows indicate the data flow directions within the system. The system consists of three main modules: The instrument placement planner, collision object converter, and robot programming interface. The instrument planner (top left block) processes user input and visualizes overlaid medical images such as MRI, as well as previews the instrument placement. The collision object converter (bottom left block), built on the motion tracking and spatial localization functionalities of MR-HMD, maps the operator's motion to a customized avatar and then converts the avatar to collision objects for robot trajectory planning. The robot programming interface (right block) is responsible for trajectory planning, preview of the planned trajectory, and waypoint adjustment. Once the operator confirms the trajectory, the robot will automatically execute the instrument placement.}\label{fig:system diagram}
\end{figure*}

Robotic-assisted medical systems (RAMS) have been rapidly developing and revolutionizing the paradigm in operating rooms in recent years. These systems are capable of enhancing dexterity, alleviating fatigue for medical staff, and minimizing errors during surgical procedures \cite{klodmann2021introduction}. These benefits are attributed to dedicated consoles that allow dexterous surgical operations and versatile imaging modalities that ensure the representation of intricate anatomy. 
%Although RAMS could assist the medical staff in labor-intensive medical operations, they still have great potential in improving the human-robot interface.

One of the core functionalities of RAMS is to position and manipulate medical instruments such as drills, milling tools, or endoscopes through a console \cite{HASSANALIDEH202034,sugano2013computer,zhou2019towards}. However, the operation of RAMS usually leads to a visual discontinuity in which the operator needs to shift focus from instruments to the screen of the navigation system, thereby detracting attention from patients \cite{qi2021holographic,skyrman2021augmented}. The absence of a line of sight to the anatomical site can increase the likelihood of medical errors such as unintended collision and sub-optimal tool placement. A constant need to switch focus would intensify the operator's fatigue and reduce efficiency during the procedure. %Additionally, the lack of mutual awareness between the robot and the operator increases the chance of accidental harm to the patient and the operator while the robot is in motion. 
Thus, it is beneficial to develop an informative and uninterrupted human-robot interface for RAMS.
% add related work to explain the advantages and applications of mixed reality in virtual overlay, robot intent communication, and robot programming
To this end, emerging mixed reality (MR) technologies shed light on potential solutions to the challenges \cite{qian2019review}.  

%Emerging MR head-mounted displays (HMDs) are able to contribute to intuitive human-robot interfaces by providing immersive visualization and multimodal sensor feedback.
The integration of MR and robotic systems in medical applications has enhanced the operator's ability in perceiving anatomical structures and surgical plans that are overlaid on real-world scenes \cite{wen2010intraoperative, Wen201231, WEN201468, InSituSpatial}.
Motivated by mitigating the gap between inexperienced operators and the required robot programming skills, researchers proposed novel MR frameworks to enable intuitive and easy-to-learn robot programming and teaching interfaces \cite{gadre2019end,ostanin2020human, Tram2023Intuitive, krieglstein2023skill,fu2023augmented}. Moreover, methodologies to convey motion intents of robots to humans have been investigated to ensure safe human-robot collaboration  \cite{rosen2020communicating, gruenefeld2020mind,bolano2021deploying}. These works commonly focus on enhancing the operator's perception and control over the robotic system, yet they overlook the importance of informing the robot about the operator's status. The integration of anatomy visualization with robot programming interfaces, coupled with human-robot mutual perception (i.e., both the operator and the robot are aware of each other), has not been fully explored in RAMS. %Aimed to improve the interface of the current RAMS and inspired by the capability of MR, 

In this paper, we propose a novel RAMS framework based on MR and human-robot mutual perception for medical instrument planning and execution. The framework enables the visualization of anatomical overlay and uses a hand-held device to facilitate the localization of anatomical targets. %Given the necessity for the robot to be aware of the operator's status and account for possible collisions, a converter is designed to transform the operator's avatar into collision objects. 
A converter is designed to transform the operator's avatar into detectable objects, allowing the robot to be aware of the operator's location and prevent possible collisions. In addition, a robot programming interface is developed for trajectory preview and waypoint adjustment.

Before implementing the proposed system, one of the indispensable prerequisites is to overlay rendered virtual objects with their physical counterparts. The accuracy of the overlay can affect user experience and the effectiveness of planning. Researchers have proposed various methods to calibrate the transformation from the virtual scene to the real world. These methods include registering two movement sets acquired by a head-mounted display (HMD) and an external tracking system \cite{elsdon2018augmented}, manually aligning virtual landmarks with their real counterparts \cite{azimi2017alignment}, and registering the point clouds of virtual model and real scene \cite{ostanin2020human}. %, and using pair correspondences detected by the MR-HMD and an external tracking system \cite{palumbo2022easy}. 
Existing methods require repetitive virtual-to-real calibration after each launch of the MR application, making the procedure time-consuming. As the amount of rendered graphics increases, running online calibration algorithms becomes even more challenging due to the limited computation resources of HMDs. %potentially limiting the development of more complex virtual scenes. 
To simplify the calibration process and enhance its reusability, we propose a virtual-to-real calibration method by combining an external tracking system and an inside-out tracking module of MR-HMDs developed by \cite{martin2023sttar}. 

To evaluate the feasibility of the proposed framework, we implemented the system in two medical use cases: coil placement in Transcranial Magnetic Stimulation (TMS) and drill and injector positioning in femoroplasty (detailed in section \ref{use cases}).

In this work, we made the following contributions:
\begin{enumerate}
    \item A novel RAMS framework that combines MR and human-robot mutual perception to provide safe medical instrument planning and execution. 
    \item An easy-to-use method for virtual-to-real calibration that facilitates the overlay of virtual scenes onto the real world.
    \item Evaluation of the proposed framework, including the virtual-to-real calibration method and the instrument alignment results using two medical use cases.
\end{enumerate} 

\section{Method}

\subsection{System Description}

The system comprises three main modules: an interface for instrument placement planning, a collision object converter based on avatar, and a robot programming interface. Together, these components help the operator observe the patient's anatomy in-situ, preview the planned trajectory of the robot, and enable secure execution of tool placement. %The system diagram is shown in Fig. \ref{fig:system diagram}.

\subsubsection{Instrument Placement Planner}
The instrument placement planner, shown in Fig.\ref{fig:system diagram}, allows the operator to determine the anatomical target of interest and intuitively preview the placement. Before confirming the target, the operator is able to observe the overlay of the patient's magnetic resonance imaging (MRI), which is preprocessed by medical image segmentation tools \cite{fedorov20123d} and 3D computer graphics software \cite{blender, meshlab}. By viewing the MRI overlay, the operator can focus on the region of interest and plan the target with a customized hand-held device (clicker). %The clicker is used for target selection and fine adjustments, relieving the clinician from frequently looking back and forth between the patient and conventional image display technologies.
The clicker, equipped with a microcontroller, an optical marker, a probe, and a joystick, can assist the operator with determining anatomical targets in-situ without shifting focus to conventional displays. A virtual ray of light is projected along the probe's axis, which intersects with the segmented medical image. On the projected area, a colored spot designates the target position in real time. After pressing the joystick, the operator can define an initial target position and subsequently fine-tune it using the joystick. The target position is then sent to the instrument pose calculator described in \cite{liu2023toward} and \cite{liu2022inside}. The preview of the instrument uses the calculated pose, and assists the operator in understanding the intended placement pose.

%This real-time, three-dimensional representation facilitates the clinician to target the anatomical spots of interest within the volumetric medical images with precision and efficiency. With an HMD, the medical images can be superimposed directly onto the patient's anatomy. %The system can also provide additional information to facilitate the identification of relevant anatomy. As an example, the system can provide annotated information on the segments of different cortical areas of the brain during TMS procedures, providing clear and labeled reference points for the clinician when pointing to the targeted anatomy using the clicker.

\subsubsection{Collision Object Converter}

Since previewing the planned pose of the instrument can help the operator identify potential sub-optimal placements that cause inconvenience or injury to the patient. Likewise, enabling the robot's awareness of the operator may help to eliminate potential collision when the robot is in motion. To achieve this, an avatar representing the operator is created using Ready Player Me \footnote{Ready Player Me, Inc., Estonia}. An HMD (Microsoft HoloLens 2) is then used to capture the motion of the operator and map that motion to the avatar. Considering the HMD's limitation in capturing full-body motion, hand-tracking and head-tracking results of the HMD are transformed, using inverse kinematics, to reconstruct the configuration of the avatar's upper limbs and head, respectively. The avatar's main components, including upper limbs, head, and main body, are then converted to simple geometric shapes and are considered as collision objects in the robot trajectory planner (see Collision Object Converter in Fig.\ref{fig:system diagram}). By converting the operator's avatar to collision objects, the framework can help the robot obtain the spatial information of the operator and avoid potential collisions. %It helps to reduce the risk of potential collisions while the clinician is performing the procedure within the working space of the robot. 

\subsubsection{Robot Programming Interface}

Once the instrument's placement pose is confirmed, the operator can send it to the robot trajectory planner, which is supported by the motion planning framework MoveIt! \cite{coleman2014reducing}. Taking the collision objects into consideration, the planner will plan a collision-free trajectory and send it back to the HMD. This step allows the operator to preview the robot's trajectory overlaid on the real world. If the current trajectory lacks in safety or precision, the operator has the option to modify the trajectory's waypoints and re-plan it to achieve better outcomes.  Once satisfied with the planned trajectory, the operator can then confirm the execution of instrument placement (see Fig.\ref{fig:system diagram}, Robot Programming Interface).

\begin{figure}[t!]
\centering 
\includegraphics[width=0.9\linewidth]{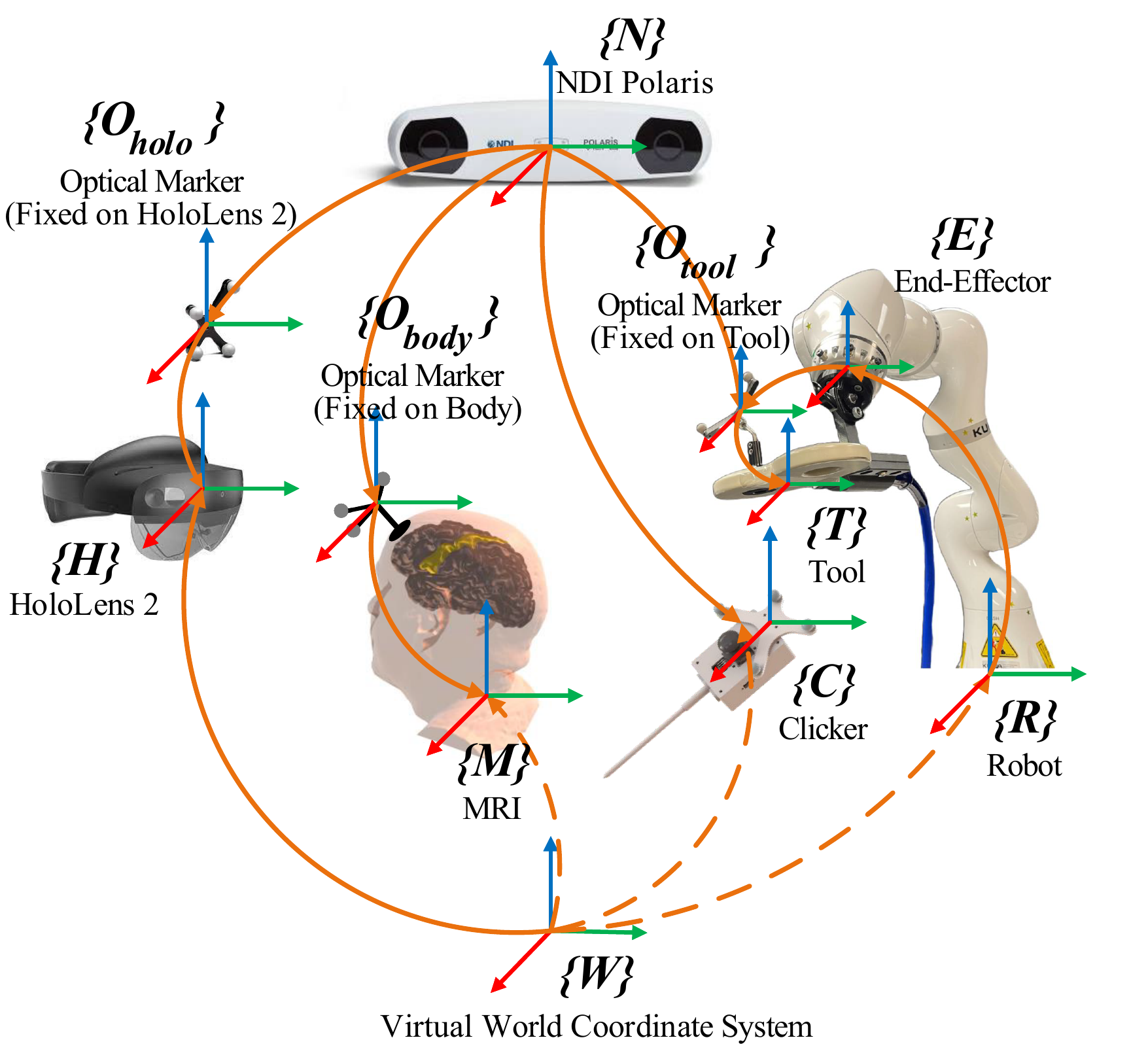}
\caption{Kinematics of the system using robot-assisted TMS as the use case. The solid arrows denote the transformations directly obtained by the external tracking system, HMD, or calibration/registration, while the dashed arrows denote the derived transformations. Note: colored RGB axes and $\{\cdot\}$ represent coordinate systems. }\label{fig:system kinamatics}
\end{figure}

\subsection{System Kinematics}

\begin{figure}[b!]
\centering 
\includegraphics[width=1\linewidth]{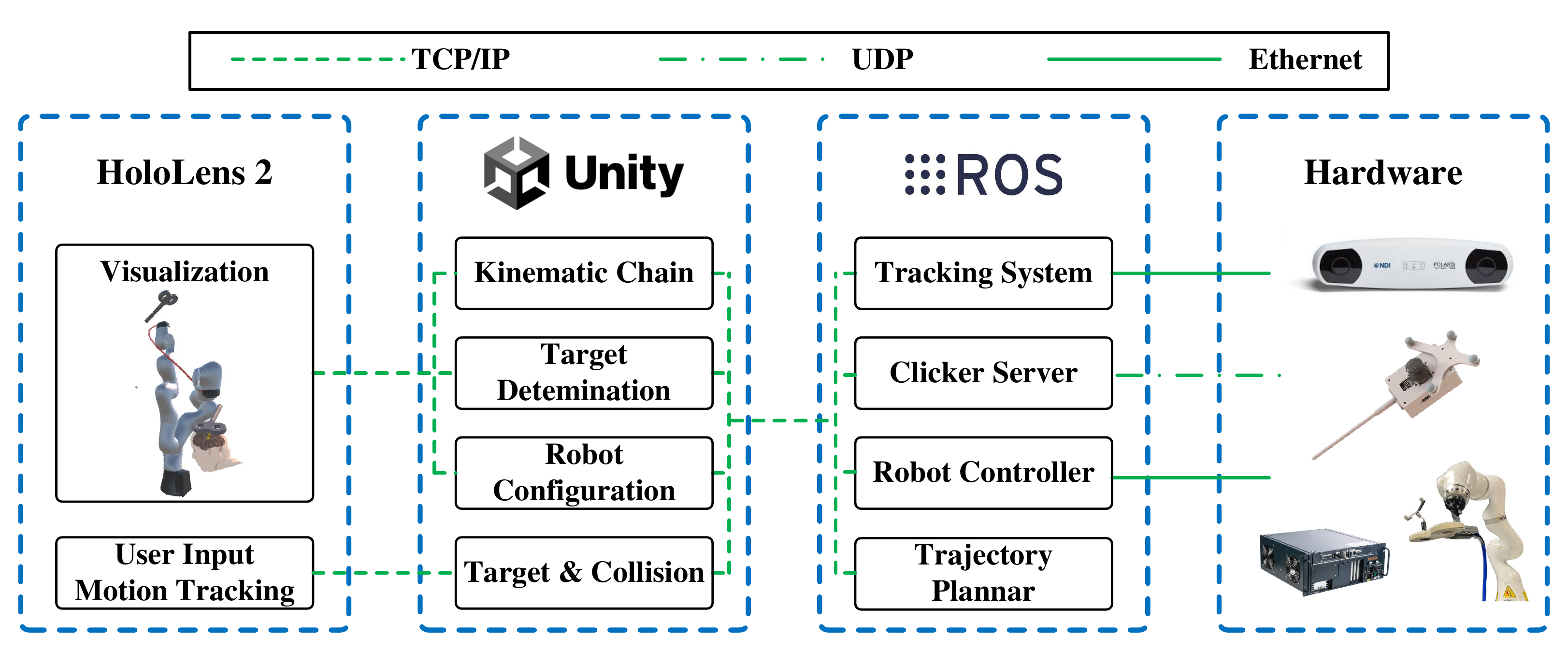}
\caption{The communication network of the system. Four main components — HoloLens 2, Unity, ROS, and Hardware — are connected using various methods. Each component encapsulates sub-elements divided by functionality. Ethernet is used for robot and sensor connections, and Transmission Control Protocol/Internet Protocol (TCP/IP) and User Datagram Protocol (UDP) are used for wireless communication.}\label{fig:system communication}
\end{figure}
Determining the placement of the medical instrument on the targeted anatomy or understanding the operator's relative position to the robot requires the kinematic transformations between each object.
%Since the system is dynamic and integrated with multiple components, it is essential to acquire their spatial poses.
An external tracking system (NDI Polaris Vicra \footnote{Polaris NDI, Northern Digital Incorporated. Ontario, Canada}) is used to track the optical markers attached to the objects. The kinematic chain of the system in robotic-assisted TMS is shown in Fig. \ref{fig:system kinamatics}. As the figure denotes, the proposed system incorporates multiple components, including an HMD (HoloLens 2), the patient's body with its MRI, a clicker, a TMS coil, and a robot.  %The solid arows denote the known kinematic transformations obtained from the external tracking system, HMD, and calibration/registration methods, while the dashed arrows represent the derived transformations based on the other known transformations. 

The transformation from the optical marker fixed on the HMD to the HMD's local coordinate system, depicted in Fig \ref{fig:system kinamatics}, is obtained through the virtual-to-real calibration described in Section \ref{sec:calibration}. The system requires subject-image registration to align the MRI with the subject's body, as well as instrument calibration to find the poses of the instrument relative to the robot and the tracking system.
%determine the transformation from the optical marker mounted on the subject's body to the coordinate system of the MRI
We register the medical image to the subject by paired-point registration and Iterative Closest Point (ICP) algorithms \cite{Arun1987,besl1992method, Maurer1997,liu2023toward}. Additionally, instrument calibration is done by using the tool center calibration method presented in \cite{liu2022inside} and the geometry-based end-effector calibration method described in \cite{liu2023GBEC}.

\subsection{System Communication}
To improve the performance of the HMD (i.e., HoloLens 2), we run the application on Unity 3D using the Holographic Remoting functionality supported by Microsoft. The communication of the proposed system (Fig. \ref{fig:system communication}) connects the HMD to a PC through TCP/IP and allows the PC to perform graphics rendering, significantly reducing the computation load of the HMD. Correspondingly, the HoloLens 2 streams back user input and motion tracking results to the applications on PC. The applications then map the motion to the avatar and converts the avatar to collision objects.  Along with the target pose of the instrument, the collision objects are sent to the trajectory planner of MoveIt!. The Robot Operating System (ROS) and the integrated medical robotic system in \cite{liu2023toward} are responsible for connecting the hardware, including the tracking system, the clicker, and the robot.  

\subsection{Virtual-to-Real Calibration}\label{sec:calibration}

\subsubsection{Calibration Method}
Accurate virtual-to-real calibration ensures the proper overlay of the rendered graphics onto physical objects. Before computing the transformation from the virtual world coordinate system $\{W\}$ to the external tracking system $\{N\}$ (shown in Fig. \ref{fig:calibration}), the proposed method first determines the transformation between the HMD's local coordinate system $\{H\}$  and the optical maker on the HMD $\{O_{holo}\}$. Besides the external tracking system (i.e., NDI Polaris), an inside-out tracking module (STTAR) deployed on HoloLens 2 is used \cite{martin2023sttar,keller2023hl2irtracking}. STTAR utilizes the built-in cameras of HoloLens 2 to localize passive retro-reflective markers and has sub-millimeter accuracy when used within 250 mm to 750 mm in depth.
%\footnote{\url{https://github.com/andreaskeller96/HoloLens2-IRTracking-Sample}}
%However, despite the benefits of better accuracy compared to image-based AR markers, stability issues still exist with STTAR. Most notably, the development mode of the HMD limits features such as holographic remoting due to compatibility constraints. The refreshing rate also drops significantly under computationally demanding AR scenes. 
Considering the larger range and better accuracy provided by NDI Polaris \cite{fattori2021technical, martin2023sttar}, the proposed system only uses STTAR in the virtual-to-real calibration as an auxiliary tool for measurement. 

%The transformation obtained by the proposed virtual-to-real calibration ensures accurate outside-in tracking. The virtual world coordinate is defined and fixed by the HMD's self-localization functionality when the application is initialized. The transformation $^{O_{holo}}_H T$ can be calculated by first finding the kinematic chain of the optical marker attached to the headset both in the virtual scene and the real world. 
The calibration scheme is shown in Fig. \ref{fig:calibration}a. We use both tracking systems to track a reference optical marker that is not attached to HoloLens 2 and then calculate the transformation $^{O_{holo}}_H T$:
\begin{equation}
    ^{O_{holo}}_{H}T = ^{N}_{O_{holo}}T^{-1} * ^{N}_{O_{ref}}T * ^{H}_{O_{ref}}T^{-1}
\end{equation}
where the transformation from HoloLens 2 to the reference optical marker $^{H}_{O_{ref}}T$ is obtained by STTAR, and the transformation matrices $^{N}_{O_{ref}}T$ and $^{N}_{O_{holo}}T^{-1}$ are obtained by NDI Polaris. Integrating $^{O_{holo}}_{H}T$ into the known kinematic chain, we can derive the transformation from the virtual world coordinate system to the NDI Polaris coordinate system:
\begin{equation}
    ^{W}_{N}T = ^{W}_{H}T*^{O_{holo}}_{H}T^{-1}*^{N}_{O_{holo}}T^{-1}
\end{equation}
With the transformation between the virtual scene and the real world, we can overlay virtual content onto real-world objects correctly. The calibration does not rely on the operator's skills. In addition, \cite{ungureanu2020hololens} has shown that the pose of the optical marker with respect to the device's origin is unchanged. Thus, the calibration result is reusable. 

\subsubsection{Calibration Evaluation}\label{calibration evaluation}
\begin{figure*}
\centering 
\includegraphics[width=0.8\linewidth]{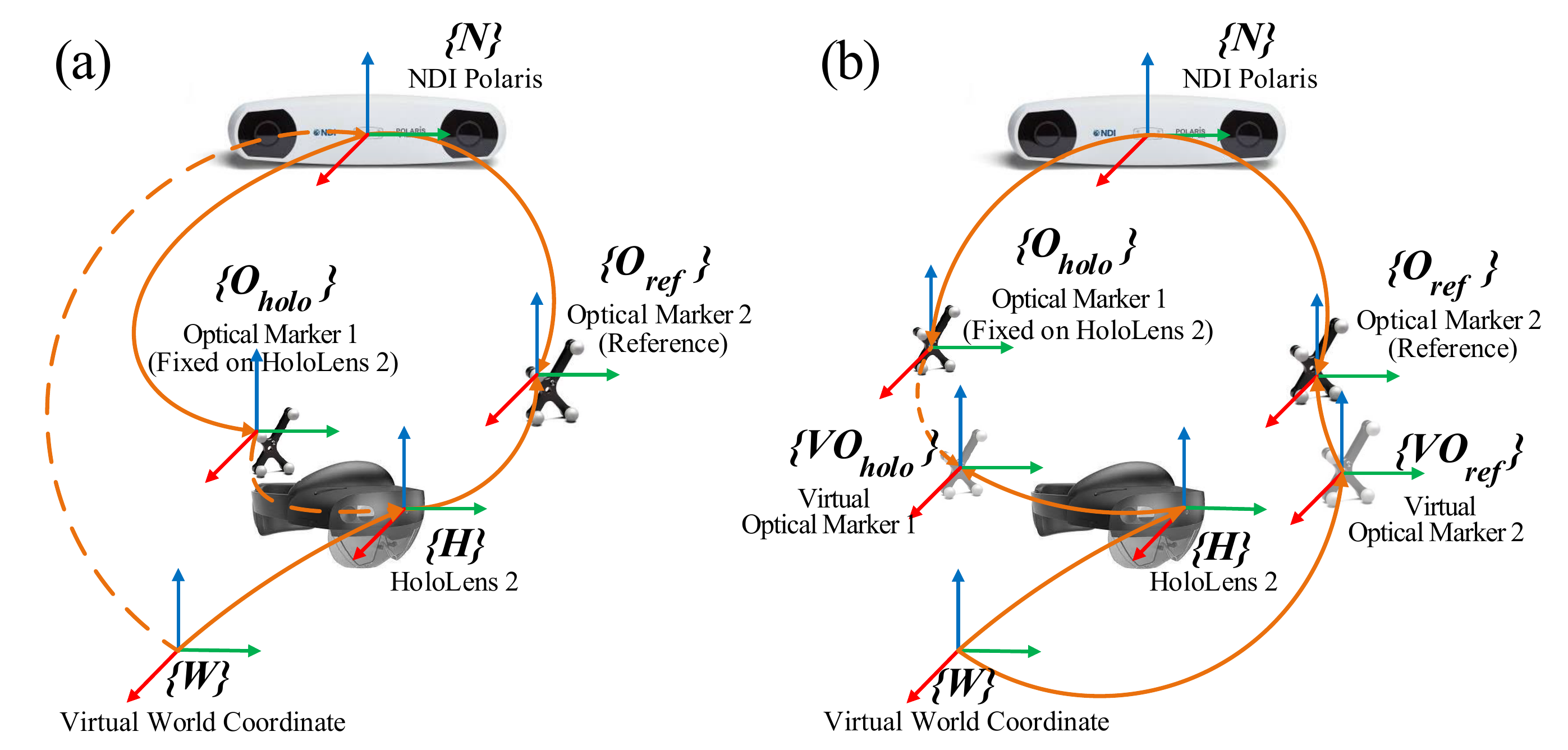}
\caption{Kinematics of the virtual-to-real calibration and its evaluation method. (a) The kinematic chain of the calibration method. Two optical markers are used: one is fixed on the HoloLens 2, and the other is free to move and serves as a reference. The transformation from HoloLens 2 to the reference marker is measured through the STTAR system \cite{martin2023sttar}. (b) The kinematic chain of the evaluation method in which the operator aligns the virtual marker $VO_{ref}$ to the real one $O_{ref}$. Note: the notations follow the same convention as outlined in Fig. \ref{fig:system kinamatics}}\label{fig:calibration}
\end{figure*}

Though the virtual-to-real calibration can be achieved without human bias, the evaluation still depends on the operator's perception. The kinematic chains of the evaluation method are shown in Fig. \ref{fig:calibration}b. In this method, we manually place the physical reference marker $O_{ref}$ in a known position and align a virtual reference marker $VO_{ref}$ on it. We can get the transformation between two real markers ($O_{ref}$ and $O_{holo}$) through NDI Polaris and the transformation between two virtual markers ($VO_{ref}$ and $VO_{holo}$) through HoloLens 2. Then, we can calculate the virtual-to-real calibration error $^{O_{holo}}_{VO_{holo}}T$ which is the transformation from the real optical marker $O_{holo}$ to the virtual optical marker $VO_{holo}$:
\begin{equation}
   ^{O_{holo}}_{VO_{holo}}T= ^{N}_{O_{holo}}T^{-1}*^{N}_{O_{ref}}T*^{VO_{ref}}_{O_{ref}}T^{-1}*^{W}_{VO_{ref}}T^{-1} * ^{W}_{H}T *^{H}_{VO_{holo}}T
\end{equation}

\begin{figure*}[h!]
\centering 
\includegraphics[width=1.0\linewidth]{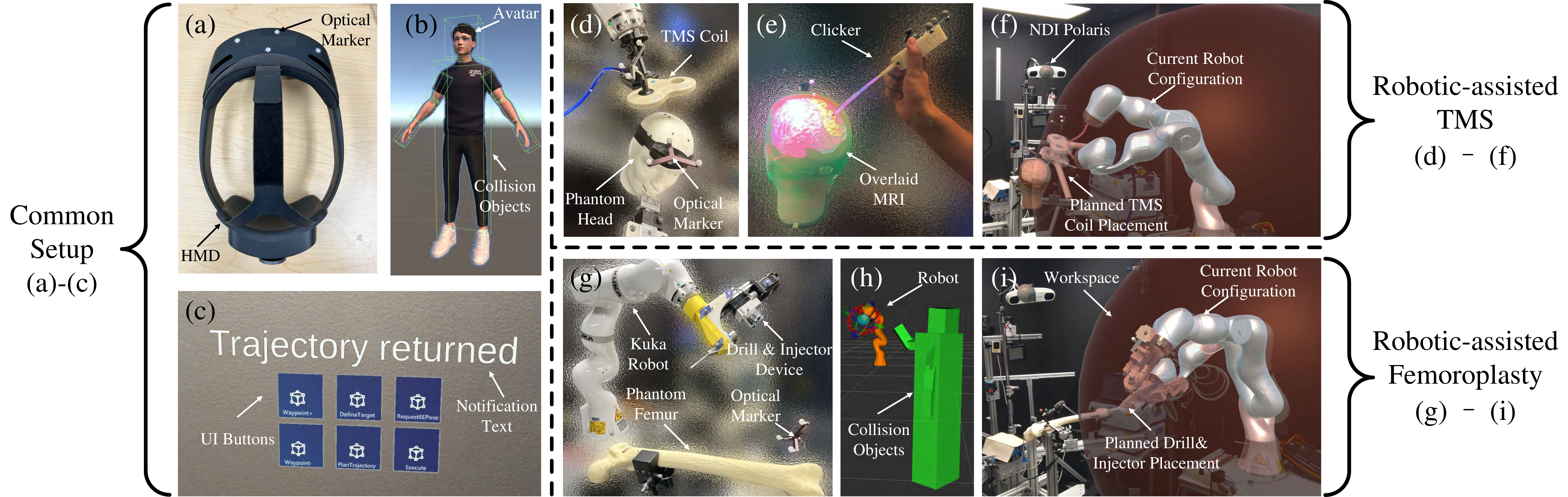}
\caption{Experimental setup. (a) HoloLens 2 with optical markers attached. (b) An avatar of the operator, with collision objects (simple geometric shapes) attached. (c) User interface. (d) TMS scene consisting of a TMS coil and a phantom head. (e) A cortical location pointed by the clicker. (f) Trajectory planning of the robot in robot-assisted TMS. (g) Femoroplasty scene with a drill and injector device, and a phantom femur. (h) Collision objects in MoveIt!. (i) Trajectory planning of the robot in robotic-assisted femoroplasty. }\label{fig:experiment}
\end{figure*}

\section{Medical Use Cases} \label{use cases}
In this section, we show the implementation details of the proposed framework in two medical cases: coil alignment in robot-assisted TMS, and drilling and injection device alignment in robot-assisted femoroplasty. 

\subsection{Robotic-Assisted TMS}
TMS is a non-invasive neuro-modulation technique that uses rapidly changing magnetic fields to stimulate a specific region of the brain. It is used in the study of brain function \cite{barker1985non,bohning1998echoplanar,chail2018transcranial} and the treatment of psychiatric and neurological disorders \cite{chail2018transcranial, DILAZZARO20212568}. TMS procedure typically involves placing a TMS coil that generates focused magnetic pulses over the head to induce micro-currents in a cortical area. Robotic-assisted TMS has shown promising advantages in improving repeatability and accuracy for TMS coil placement and avoiding operator's fatigue by eliminating the need for manual alignment \cite{harquel2017automatized,noccaro2021development,liu2022inside,liu2023toward}. Nevertheless, novel interaction and visualization methods in the context of robotic-assisted TMS have not been explored. In this work, we assess the feasibility and the instrument placement accuracy of the proposed MR framework deployed on robotic-assisted TMS.

TMS coil placement is an alignment task that moves the robot-held coil to a planned pose by commanding the robot. %The primary objective of the alignment is to position and orient the TMS coil on the head of the subject at a planned location and orientation determined from MRI. The alignment task requires an optical tracking camera (NDI Polaris), an optical marker attached to the TMS coil, and another optical marker affixed to the head of the subject. 
The planned pose of the TMS coil represents the desired pose on the scalp corresponding to a target cortical location in the brain. %Fig. \ref{fig:system kinamatics} represents the kinematic chain associated with this task.  %Interfaced with the neuronavigation and robotic system in \cite{liu2023toward}, 
Accordingly, the MR framework for robot-assisted TMS has the following workflow:
\begin{enumerate}
    \item The operator registers the MRI to the subject's head and calibrates the TMS coil with the robot. 
    \item By using the clicker, the operator determines a cortical location of interest. The target location is then fed to the neuronavigation system detailed in \cite{liu2022inside, liu2023toward} for calculating the TMS coil pose. 
    \item The preview of the coil placement is presented to the operator who can either start the robot trajectory planning or redefine the target location by using the clicker.
    \item Incorporating the spatial information of the operator, the trajectory planner of MoveIt! takes the potential collision into consideration and generates a feasible trajectory.
    \item By observing the trajectory overlaid onto the real world, the operator has the option to redefine waypoints to create new trajectories or proceed with the placement if it aligns with expectations.
\end{enumerate}

\subsection{Robotic-Assisted Femoroplasty}
Femoroplasty, or \textit{femoral bone augmentation}, is a minimally invasive surgical procedure designed to prevent hip fractures \cite{beckmann2007femoroplasty,sutter2010biomechanical}. In femoroplasty,
bone cement is injected into the proximal femur to increase its strength \cite{beckmann2011fracture,farvardin2019significance}. Previous works have explored surgical trajectory planning and execution in femoroplasty and developed robotic methods to improve drilling and injection accuracy \cite{basafa2015subject,farvardin2021biomechanically,bakhtiarinejad2023surgical}. 

The main task of robot-assisted femoroplasty is to align a custom-designed Robotic Drilling and Injection Device (RDID) to the line connecting the greater trochanter surface of the femur and the injection point \cite{bakhtiarinejad2023surgical}. The use case of robot-assisted femoroplasty shares the same kinematics shown in Fig. \ref{fig:system kinamatics}. The workflow of the femoroplasty adheres to the same policy of the TMS case and also involves subject-image registration, anatomical target planning, instrument pose computation, robot trajectory planning, and placement execution. 

\section{Experiments}
\subsection{Experiment Setup}
The proposed system operates on two host PCs. PC 1 is equipped with an Intel Core i7-12700H Processor, 16 GB DDR5 RAM, and an NVIDIA GeForce RTX 3060 Laptop GPU. It runs Unity 2021.3.8f1 with the Mixed Reality Toolkit (MRTK) 2.8.2 on Windows 10 and connects with a Microsoft HoloLens 2 via the Holographic Remoting Player 2.9.1. Meanwhile, PC 2 runs ROS Ubuntu 20.04 and interfaces with a robotic arm (LBR iiwa 7 R800 robot, Kuka, Germany) and an optical tracking system (NDI Polaris Vicra, Northern Digital Inc., Canada). A microcontroller (Nano 33 IoT, Arduino, Italy) embedded in the clicker communicates with the ROS network via UDP. The experiment setup can be viewed in Fig. \ref{fig:experiment}.
\subsection{Experiment Results}
\subsubsection{Evaluation of Virtual-to-Real Calibration}
\begin{figure}[h!]
\centering 
\includegraphics[width=1.0\linewidth]{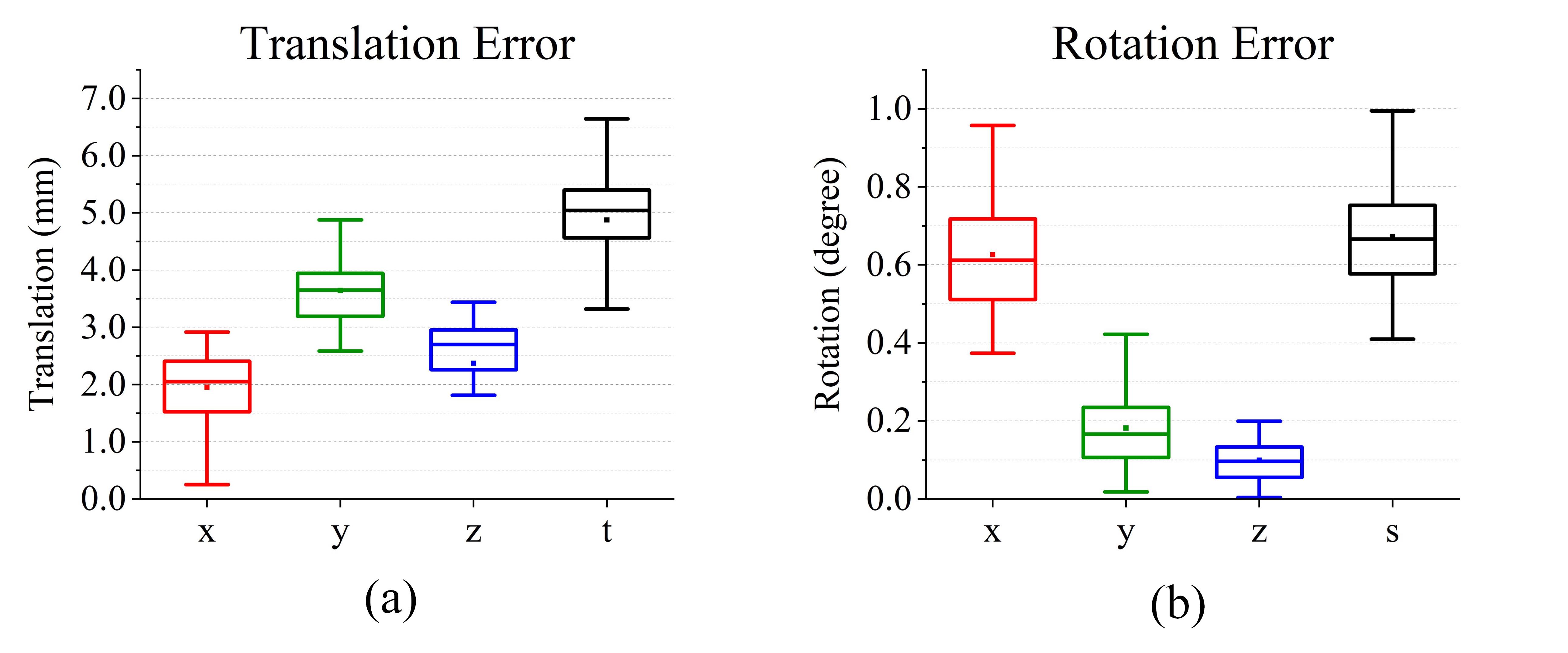}
\caption{Calibration errors of the transformation between the local coordinate system of the HMD and the optical marker affixed on the HMD. The results show the errors in 50 trails of calibrations. The errors are separated into translation and rotation about the x, y, and z axes. The overall translation error is denoted by t in (a), and the overall rotation error around a single rotation axis is denoted by s in (b).}\label{fig:Calibration Error}
\end{figure}

For error analysis, we conducted 50 trials of virtual-to-real calibrations in which the operator held the HMD and placed the reference optical marker in different poses each time. After obtaining the calibration results, the operator performed the evaluation by aligning a virtual reference marker, using hand gestures and a keyboard, to the real marker in a single trial. Fig. \ref{fig:Calibration Error} shows both translation and rotation errors of the calibration error defined in section \ref{calibration evaluation}. %The translation errors are $x = 2.05\pm 0.44 $ mm, $y = 3.65 \pm 0.38$ mm, and $z = 2.70 \pm 0.35 $ mm, respectively, while the rotation errors are  $\theta_x = 0.61\pm 0.10 $\degree, $\theta_y = 0.17 \pm 0.06 $\degree, and $\theta_z = 0.10 \pm 0.04 $\degree, respectively. 
The overall translation error (Euclidean) is $4.88\pm 0.85 $ mm, and the rotation error (angle-axis representation) is  $0.67\pm 0.14$\degree.

\subsubsection{Evaluation of Medical Cases}
\begin{figure}[h!]
\centering 
\includegraphics[width=1\linewidth]{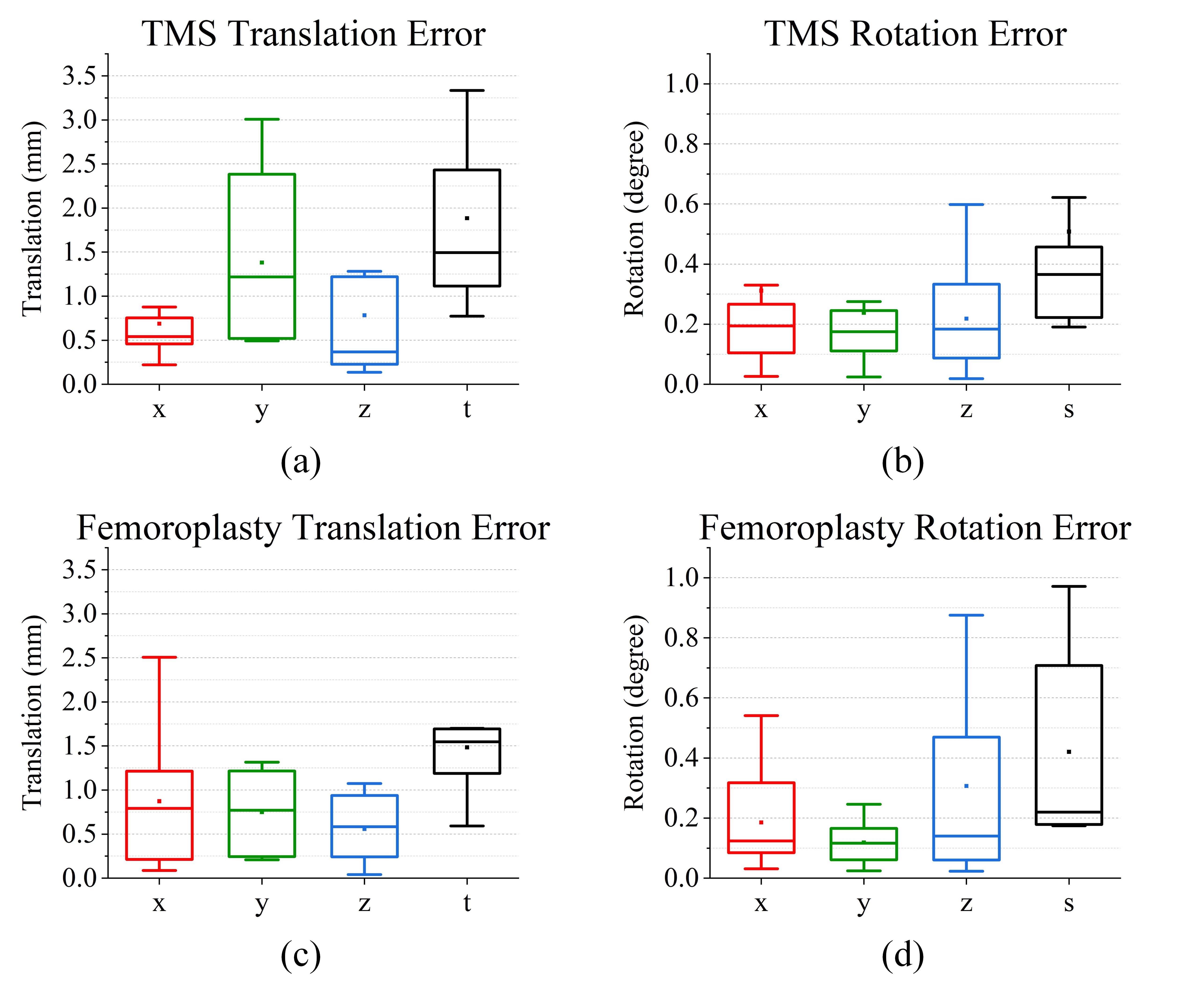}
\caption{Medical instrument placement results. (a) and (b) present the translation and rotation errors of the coil placement in TMS, respectively. (c) and (d) present the translation and rotation errors of the RDID placement in femoroplasty, respectively. The notations follow the same convention described in Fig \ref{fig:Calibration Error}. }\label{fig:Experiment results}
\end{figure}

We conducted 10 trials of different alignments for each medical case. In each trial, we use the proposed framework to plan a target pose on the body (phantom head for TMS and phantom femur for femoroplasty) and command the robot to execute the alignment. Fig. \ref{fig:Experiment results} shows the alignment errors between the planned and executed target poses. The average translation errors of TMS and femoroplasty are below 2 mm ($1.88\pm 1.21 $ mm and $1.48\pm 0.53 $ mm, respectively), while the average rotation errors are below $0.6\degree$ ($0.51 \pm 0.51\degree$ and $0.42 \pm 0.31\degree$, respectively). The results show the viability and accuracy of the system in both medical cases.  

%Robotic tool alignment results. (a) and (c) present the translation errors of the coil placement in TMS and the RDID placement in femoroplasty, respectively. The translation errors are calculated along the x y, and z axes. The overall Euclidean errors are denoted by t. (b) and (d) present the rotation errors of the instrument alignment in two applications. The overall rotation error is denoted by s

\section{Discussion}
Compared with the sub-millimeter accuracy of STTAR and NDI Polaris, the calibration errors in translation were relatively large. The issue may arise from the insufficient alignment skills and perception bias of the operator since a small rotation difference in the alignment could cause a large translational offset. An user study that involves multi-users and multi-trials may better reflect the quality of the calibration.
The calibration process requires the operator to hold the HMD and collect the data. Thus, the other error sources may include the tremor of the operator and the data asynchrony between the HMD and the tracking system. The translation accuracy might be further improved with an auxiliary stand for stabilizing the HMD. 
 
The calibration error affects the accuracy in determining the anatomical target. However, it does not affect the kinematic accuracy of the robot execution which depends on subject-image registration and instrument calibration instead. The decoupled relationship is supported by the low translation and rotation errors in the medical cases compared to the calibration error. The calibration result also affects collision avoidance which needs the pose of the operator. To compensate for the calibration error, the operator could set a larger volume for each geometric shape converted from the avatar. 
 
Since the current work only presents a technical pilot study, a detailed user study is required to thoroughly validate the robustness and efficacy of the proposed framework. This study should involve recruiting participants to contrast our method with the traditional RAMS. A more comprehensive data analysis could be carried out after obtaining sufficient user samples.  Additionally,
further improvements are worth exploring such as developing multimodal input and feedback for the hand-held device and enhancing the robustness of the trajectory planning process that considers the dynamic nature of human motions.

%the operator's spatial pose with respect to the robot while the kinematic accuracy of the robot execution still depends on NDI Polaris, which has a high accuracy of $\approx0.25$ mm \cite{koivukangas2013technical}, and the collision objects derived from the avatar could be set with adequate tolerance.

\section{Conclusion}
In this work, we propose a novel RAMS framework using MR to facilitate the planning and execution of medical instrument placement. The framework provides informative image overlays and a handheld device for the identification of anatomical targets. Additionally, the proposed system ensures safe human-robot interaction through mutual perception. An easy-to-implement calibration technique is also presented to align virtual content and the real world accurately. The system was verified through its application in two medical scenarios: robot-assisted TMS and femoroplasty. The results demonstrate the system's practical viability and its potential for broader application in various RAMS contexts.
%In this work, we propose a novel RAMS framework using MR to facilitate medical instrument planning and execution. Besides providing informative image overlays and a hand-held device to readily determine the anatomical target, it realizes a safe human-robot interaction based on human-robot mutual perception. To accurately overlay the virtual content with the real world, an easy-to-implement calibration method is proposed.  In the end, the system was verified through two medical cases: robot-assisted TMS and femoroplasty. The results demonstrate the system's favorable feasibility and its potential applicability to other RAMS. 

% that not only allows the operator to preview the robot's intended motions but also makes the robot aware of the operator and avoids potential collisions

 %It provides features that enable informative medical imaging visualization, and a user-friendly hand-held input device (clicker) assists in the planning of the instrument placement. In addition, it provides an effective robot programming interface that allows the operator to readily define waypoints of the instrument. The trajectory planner also incorporates the spatial information of the operator, reducing the risks of collision between the operator and the robot. 

% \addtolength{\textheight}{-12cm}  

\bibliographystyle{IEEEtran}
% \clearpage
\bibliography{ref}

\end{document}